\documentclass{article}

     \usepackage[final]{neurips_2020}

\usepackage[utf8]{inputenc} 
\usepackage[T1]{fontenc}    
\usepackage{hyperref}       
\usepackage{url}            
\usepackage{booktabs}       
\usepackage{amsfonts}       
\usepackage{nicefrac}       
\usepackage{microtype}      
\usepackage{graphicx}
\usepackage{amsmath}
\usepackage{amssymb}
\usepackage{natbib}
\usepackage{subfig}
\UseRawInputEncoding
\usepackage{float}

\title{One Shot Audio to Animated Video Generation}

\author{%
  Neeraj Kumar \\
  Hike Private Limited\\
  \texttt{neerajku@hike.in}\\
  \And
  Srishti Goel \\
  Hike Private Limited \\
  \texttt{srishtig@hike.in}\\
  \AND
  Ankur Narang\\
  Hike Private Limited \\
  \texttt{ankur@hike.in} \\
  \And
  Brejesh Lall \\
  IIT Delhi \\
  \texttt{brejesh@ee.iitd.ac.in} \\
  \And
  Mujtaba Hasan \\
  Hike Private Limited  \\
  \texttt{mujtaba@hike.in}
  
  \And
  Pranshu Agarwal \\
  Hike Private Limited  \\
  \texttt{pranshu@hike.in}
  
  \And
  Dipankar Sarkar \\
  Hike Private Limited  \\
  \texttt{dipankars@hike.in}
}

\begin{document}

\maketitle

\begin{abstract}
    We consider the challenging problem of audio to animated video generation. We propose a novel method \textit{OneShotAu2AV} to generate an animated video of arbitrary length using an audio clip and a single unseen image of a person as an input. The proposed method consists of two stages. In the first stage, OneShotAu2AV generates the talking-head video in the human domain given an audio and a person$'$s image. In the second stage, the talking-head video from the human domain is converted to the animated domain.
The model architecture of the first stage consists of spatially adaptive normalization based multi-level generator and multiple multi-level discriminators along with multiple adversarial and non-adversarial losses. The second stage leverages attention based normalization driven GAN architecture along with temporal predictor based recycle loss and blink loss coupled with lip-sync loss, for unsupervised generation of animated video. In our approach, the input audio clip is not restricted to any specific language, which gives the method multilingual applicability. OneShotAu2AV can generate animated videos that have: $(a)$ lip movements that are in sync with the audio, $(b)$ natural facial expressions such as blinks and eyebrow movements, $(c)$ head movements. Experimental evaluation demonstrates superior performance of OneShotAu2AV as compared to U-GAT-IT and RecycleGan on multiple quantitative metrics including KID(Kernel Inception Distance), Word error rate, blinks/sec.
\end{abstract}

\section{Introduction}
Audio to Video generation has numerous applications across industry verticals including film making, multi-media, marketing, education and others. In the film industry, it can help through automatic generation from the voice acting and also occluded parts of the face. Additionally, it can help in limited bandwidth visual communication by using audio to auto-generate the entire visual content or by filling in dropped frames. High-quality video generation with expressive facial movements is a challenging problem that involves complex learning steps.
Most of the work in this field has been centered towards the mapping of audio features (MFCCs, phonemes) to visual features (Facial landmarks, visemes etc.)~\citep{Alpher04,audio2face,visemes1,visemes2}. Further computer graphics techniques select frames of a specific person from the database to generate expressive faces. Few techniques which attempt to generate the video using raw audio focus for the reconstruction of the mouth area only~\citep{Chung17b}. Due to a complete focus on lip-syncing, the aim of capturing human expression is ignored. Further, such methods lack smooth transition between frames which does not make the final video look natural. Regardless, of which approach we use, the methods described above are either subject dependent~\citep{article12,temporalLoss} or generate unnatural videos~\citep{Wiles18}  due to lack of smooth transition and/or require high compute time to generate video for a new unseen speaker image for ensuring high-quality output~\citep{NeuralHead1}.Recent unsupervised approaches such as RecycleGAN~\citep{Recycle-GAN} and U-GAT-IT~\citep{ugatit} either generate low quality videos or has low expressives due to lack of eye blinks, eyebrow movement and head movement.We propose a novel approach based on two novel stages that convert an audio and single image of a person to an animated video. The first stage generates a speaker-independent and language-independent high-quality natural-looking talking head video from a single unseen image and an audio clip. It captures the word embeddings from the audio clip using a pretrained deepspeech2 model\citep{DeepSPeech2} trained on Librispeech corpus\citep{librispeech}. These embeddings and the image are then fed to the multi-level generator network which is based on the Spatially-Adaptive Normalization architecture~\citep{park2019SPADE}. Multiple multi-level discriminators~\citep{wang2018pix2pixHD} are used to ensure synchronized and realistic human video generation.
A multi-level temporal discriminator is modeled which ensures temporal smoothening along with spatial consistency. Finally, to ensure lip synchronization we use SyncNet architecture~\citep{assael2016lipnet} based discriminator applied to the lower half of the image. To make the generator input-time independent, a sliding window approach is used. Since, the generator needs to finally learn to generate multiple facial component movements along with high video quality, multiple loss functions both adversarial and non-adversarial are used in a curriculum learning fashion. For fast low-cost adaptation to an unseen image, a few output update epochs suffice to provide \textbf{\textit{one-shot}} learning capability to our approach. 
    The second stage couples an attention-based normalization driven GAN architecture with temporal predictor based recycle loss and blink loss and lip-sync loss to generate high quality animated video from human video obtained from the first stage.
    
Specifically, we make the following contributions:\

$(a)$ We present a novel approach, $\bf{OneShotAu2AV}$, that uses two stages, that are trained independently, to convert audio and single image input to animated video.\\
$(b)$ The first stage takes audio and single image of a person as input, and, leverages curriculum learning to simultaneously learn movements of expressive facial components and generate a high-quality talking-head video of the given person. The stage feeds the features generated from the audio input directly into a generative adversarial network and it adapts to any given unseen selfie by applying one-shot learning with only a few output update epochs.\\
$(c)$ The second stage leverages attention based normalization driven GAN architecture along with temporal predictor based recycle loss and blink loss coupled with lip-sync loss, for unsupervised generation of animated video that demonstrates eye blinks, eyebrow movements and lip-synchronization with audio.\\
$(d)$ Experimental evaluation demonstrates superior performance of OneShotAu2AV as compared to U-GAT-IT and RecycleGan on multiple quantitative metrics including KID(Kernel Inception Distance), Word error rate, blinks/sec.
\section{Related Work}
A lot of work has been done in synthesizing realistic videos in the human domain or animated domain from an audio clip and an image as an input. Speech is a combination of content and expression and there is a perceptual variability of speech that exists in the form of various languages, dialects, and accents.
\paragraph{Audio to Human Domain Video Generation} The earliest methods for generating videos relied on Hidden Markov Models which captured the dynamics of audio and video sequences. Simons and Cox\citep{SimonCox} used the Viterbi algorithm to calculate the most likely sequence of mouth shape given the particular utterances. Such methods are not capable of generating high-quality videos and lacked emotions. \par
CNN based models have been used to generate realistic videos given audio and single image as an input. Audio2Face~\citep{audio2face} model uses the CNN method to generate an image from audio signals. ~\citep{Chung17b}(Speech2Vid) has used an encoder-decoder based approach for generating realistic videos. Other approaches such as Synthesizing Obama: learning lip sync~\citep{article12} from the audio are trained for a single image. LumiereNet~\citep{temporalLoss} uses LSTM, DensePose~\citep{densepose} and Pix2Pix~\citep{Pix2Pix} for generating videos. These however, have limitations either in terms of expressions such as lip-sync, eye-blink, emotions or they are specifically trained for a single image and are not generalizable. We propose a spatially adaptive generator along with multiple discriminators which generates high-quality, lip-synchronized video with expressions such as eye-blink, etc.\par
~\citep{NeuralHead1} uses meta-learning to create videos of unseen images. Few shot Video to Video Synthesis~\citep{wang2019fewshotvid2vid} is able to generate videos on unseen images given a video as an input by using a network weight generation module for extracting the pattern. Such a method is computationally expensive compared to our approach which is a one-shot approach for video generation. Realistic Speech-Driven Facial Animation with GANs (RSDGAN) ~\citep{Alpher05} uses a GAN based approach to produce quality videos. They have used identity encoder, context encoder and frame decoder to generate images and used various discriminators to take care of different aspects of video generation. They have used frame discriminator to distinguish real and fake images, sequence discriminator to distinguish real and fake videos, and synchronization discriminator for better lip synchronization in videos. We introduce spatially adaptive normalization along with a one-shot approach and implemented curriculum learning to produce better results. This is explained in Sections 3,4. 

\paragraph{Video to Animated Video generation} Initially, phonemes and visemes based methods were used to create the stylized characters. ~\citep{Alpher04} has used an LSTM based approach to generate live lip synchronization on a 2D animated character. Some of these methods target rigged 3D characters or meshes with predefined mouth blend shapes that correspond to speech sounds ~\citep{inproceedings1,article09,inproceedings3,article10,article11,article12}, while others generate 2D motion trajectories that can be used to deform facial images to produce continuous mouth motions ~\citep{2dpaper,2dpaper1}. These methods are primarily focused on mouth motions only and do not show emotions such as eye-blink, eye-brow movements, etc. \par
Several works have been done on expression and facial action units classification and mapping it to the animated version of a person such as ~\citep{lucia, avatar, facshuman, greta}. They cover a finite space in terms of expression, movements and are not personalized to specific people. The proposed model is able to capture these various aspects such as facial expressions, lip-syncing, eye-blinks etc., due to attention-based generator and discriminator. \par
~\citep{highfidelavatar} uses facial action coding system~\citep{facs}, ensures lip syncing using phoneme classifier~\citep{phoneme}, expression control using ~\citep{emacs} and bone control units to create an avatar. They use an unreal engine~\citep{unreal} for generating the avatar. They use a classification-based approach to create an avatar which covers the finite space in terms of facial details in a video. On the other hand, the proposed method uses an unsupervised generative approach to create animated videos with various facial details. \par
The recent introduction of GAN~\citep{Authors14b} has shifted the focus of the machine learning community towards generative models. Several works have been done in the image to image translation as well as video to video translation. Techniques such as Pix2Pix~\citep{Pix2Pix}, Pix2PixHD~\citep{Pix2pixhd}, SPADE~\citep{park2019SPADE} work in image to image translation, but require a paired form of training. CycleGan~\citep{cyclegan} which uses cycle consistency loss, deals with unpaired form of training but lack in preserving temporal information while generating the animated videos.\par
For Video to Video style transfer, RecycleGan~\citep{Recycle-GAN} uses unpaired but ordered streams of data for both domains. This method uses recycle loss apart from adversarial and cycle loss to handle temporal information. Due to the Unet~\citep{Unet} generator and lack of attention-based architecture, it is not able to generate high-quality animated video. The proposed method uses adaptive layer and instance normalization (AdaLin) and attention-based networks along with temporal discriminator which give better superior quality animated videos over RecycleGan.\par
U-GAT-IT~\citep{ugatit} uses AdaLin and attention maps for translating an image from one domain to another. However, this architecture is not able to capture the temporal information and lacks lip synchronization and expressions such as eye-blink, eyebrow movement as well as head movement. We have leveraged AdaLin and attention map based architecture along with temporal predictor using recycle loss, blink loss and lip-sync loss for the better expression capture in animated videos(refer Section 5) in unsupervised fashion. OneShotAu2AV is able to synthesize a personalized animated video from an audio clip and a single image of the person.

\section{Architectural Design}

OneShotAu2AV consists of 2 stages: Stage 1 to generate realistic human domain videos given an audio and single unseen image as an input and  Stage 2 to generate animated videos from realisitic human domain generated videos.

\begin{figure}[h!]
    \centering
    \subfloat{{\includegraphics[width=6cm]{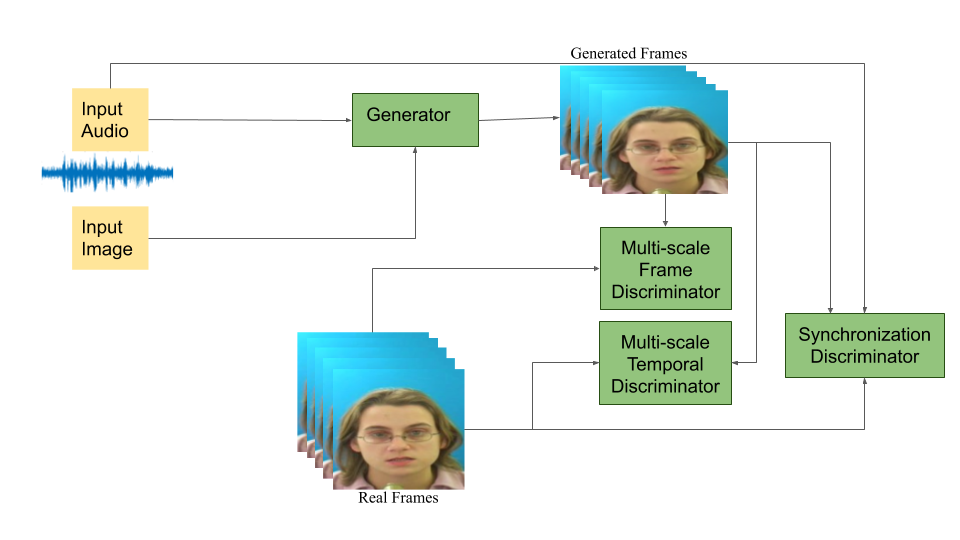} }}
    \qquad
    \subfloat{{\includegraphics[width=6cm]{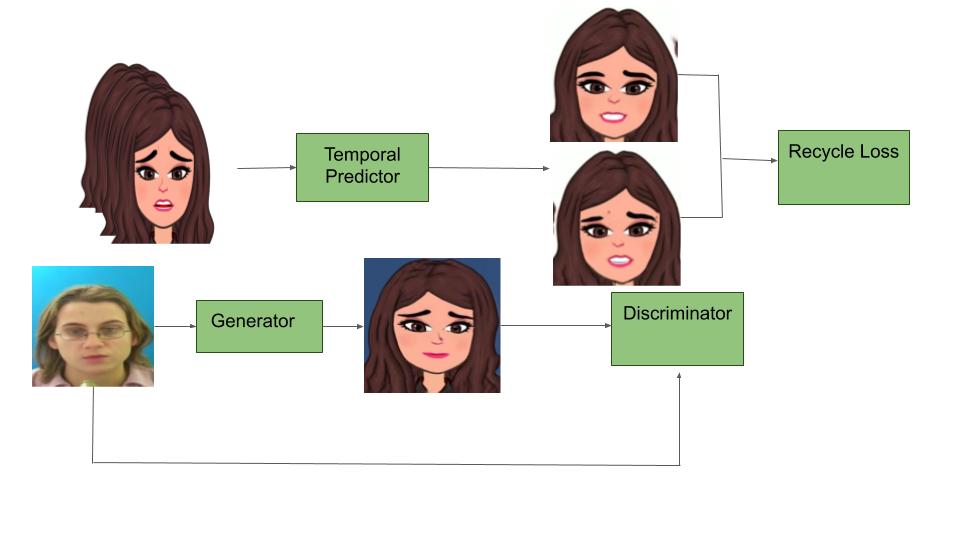} }}
    \caption{(a) Left side: Stage 1 of OneShotAu2AV with a generator and three discriminators for generating human-domain video. (b) Right side: Stage 2 of OneShotAu2AV with a generator, temporal predictor and a discriminator for generating a high quality animated video.}
    \label{fig:arch}
\end{figure}

\subsection{Stage 1}
It consists of a single generator and 3 discriminators as shown in Figure ~\ref{fig:arch}(a).
\subsubsection{Generator}
\paragraph{Spatially Adaptive Generator:}
The initial layers of generator, G uses deepspeech2~\citep{DeepSPeech2} layers followed by Spatially-Adaptive normalization similar to SPADE architecture ~\citep{park2019SPADE}. Instead of using semantic map as the input, we use the real image as an input to the SPADE generator. This helps in minimizing the loss of information due to normalization. 

\paragraph{Audio features using deepspeech2 model:}
The MFCC coefficients of audio signals are fed to the deepspeech2 model to extract the content related information of audio. The initial few layers are being used which goes to the generator .This helps in achieving  better lip synchronization in video and lower word error rate

\subsubsection{Discriminator}
We have used 3 discriminators namely a multi-scale frame discriminator, a multi-scale temporal discriminator and a synchronization discriminator.

\paragraph{Multi-scale Frame Discriminator}

 Multi-scale discriminator~\citep{wang2018pix2pixHD}, D is used in the proposed model to distinguish the coarser and finer details between real and fake images. Adversarial training with the discriminator helps in generating realistic frames. To have high resolution generated frames, we need to have an architecture with better receptive field.  A deeper network can cause overfitting, to avoid that, multi-scale discriminators are used. 

 \paragraph{Multi-scale Temporal Discriminator}

Every frame in a video is dependent on its previous frames. To capture the temporal property along with a spatial one, we have used a multi-scale temporal discriminator~\citep{temporalLoss}. This discriminator is modeled to ensure a smooth transition between consecutive frames and achieve a natural-looking video sequence. The multi-scale temporal discriminator is described as

\begin{equation}
    L(T,G,D) = \sum_{i=t-L}^{t}[\log(D(x_{i}))] + [\log(D(1-G(z_{i})))]
\end{equation}

where t is the time instance of an audio and L is the length of the time interval for which the adversarial loss is computed. 

\paragraph{Synchronization Discriminator}

To have coherent lip synchronization, the proposed model uses SyncNet architecture proposed in Lip Sync in the wild~\citep{Chung16a}. The input to the discriminator is an audio signal of 200ms time interval(5 audio signals of 40ms each) and 5 frames of the video. The lower half of the frame of resized to (224,224,3) is fed as an input.

\subsubsection {Losses}

We have used adversarial loss, L\textsubscript{GAN}, temporal adversarial loss, L(T,G,D), feature loss, L\textsubscript{FL}, reconstruction loss, L\textsubscript{RL}, perceptual loss, L\textsubscript{PL}, contrastive loss, L\textsubscript{CL} and blink loss, L\textsubscript{BL} to generate high quality output. Objective function is given below and detailed explanation is given in supplementary material.

\begin{multline*}
    min_{G}((max_{D1,D2,D3}\sum_{k=1,2,3}^{}L_{GAN}(G,D_{k})+ max_{D1,D2,D3}\sum_{k=1,2,3}^{}L(T,G,D)) + \\ \lambda_{FM} \sum_{k=1,2,3}^{}L_{FM} + \lambda_{PL}L_{PL} + \lambda_{CL}L_{CL}+\lambda_{BL}L_{BL})
\end{multline*}

where $\lambda_{FM} , \lambda_{PL},\lambda_{CL},\lambda_{BL}$ are the hyperparameters that control the importance of various loss functions in the above objective function

\paragraph{Curriculum Learning}

We have trained the model in multiple phases so that it can produce better results. In the first phase we have used a multi-scale frame discriminator and applied the adversarial loss, feature matching loss and perceptual loss to learn the higher-level features of the image. When these losses stabilize, we move to the second phase in which we have added a multi-scale temporal discriminator and synchronization discriminator and used reconstruction loss, Contrastive loss and temporal adversarial loss to get a better quality image near mouth region and coherent lip synchronized high-quality videos. After the stabilization of the above losses, we have added blink loss in the third phase to generate a more realistic image capturing emotions such as eye movement and eye blinks. 

\paragraph{Few shot learning}

To achieve a more sharp and a better image quality for an unseen subject, we have used one shot approach using perceptual loss during inference time. Our approach is computationally less expensive as compared to ~\citep{NeuralHead1,wang2019fewshotvid2vid} which we have described in Section 2 and because of the spatially adaptive nature of generator architecture, we are able to achieve high-quality video. We run the model for 5 epochs during inference time to get high-quality video frames.

\subsection{Stage 2}

The stage 2 as shown in Figure~\ref{fig:arch}(b) consists of a generator, a temporal predictor and a discriminator to generate high-quality animated videos.

\subsubsection{Generator}
The Generator model $G_{s->t}$ consists of encoder $E_{s}$, attention-based normalisation driven decoder $D_{s}$.The attention-based adaptive instance and layer normalization (AdaLin) is inspired by Class Activation Map(CAM)~\citep{CAM} which is trained to learn weights of feature maps of the source domain using the global average pooling and global max pooling. This helps the generator to focus on the source image regions that are more discriminative from the target domain, such as eyes and mouth. AdaLin adjusts the ratio of IN and LN in the decoder according to source and target domain distributions to have the features of the source domain as well as the style of the target domain.

\subsubsection{Discriminator}
The discriminator, $D_{t}$ consists of Encoder, $E_{D_{t}}$ and auxiliary classsifier, $n_{D_{t}}$. The discriminator concentrates its attention to determine whether the target image is real or fake by visualizing local and global attention maps so that it helps the generator to capture the global structure (e.g., face area and near of eyes) as well as the local regions.

\subsubsection{Temporal Predictor}
We use unpaired data but have ordered streams of frames, $(x_{1} \cdots x_{n})$ and $(y_{1} 
\cdots y_{n})$ for source and target domains. To learn better mapping from source to target domain, we focus to learn the temporal information. We introduce a temporal predictor($P_{x}$) whose architecture is same as of UNet~\citep{Unet}  which predicts the future frames given past frames as an input. This is trained with L2 loss. 

\subsubsection{Losses}

The losses such as adversarial loss, $L_{GAN}(G,D)$, identity loss, CAM loss, $L_{cam}$  are used for the domain transfer and recycle loss, $L_{recycle}$, lip sync loss, $L_{lip}$ and blink loss, $L_{BL}$ to extract the spatial and temporal information from a video that helps in generating high-quality expressive animated videos.Detailed explanation is given in supplementary material. Objective function is given below:

\begin{multline*}
    min_{G}((max_{D}L_{GAN}(G,D)+\lambda_{cam} L_{cam}^{D_{t}})  + \lambda_{recycle}L_{recycle}(G_{x},G_{y},P_{y})\\
+\lambda_{identity}L_{identity} +\lambda_{cam} L_{cam}^{s->t} +\lambda_{lip} L_{lip} + \lambda_{BL}L_{BL})
\end{multline*}

where $\lambda_{cam}$ , $\lambda_{recycle}$,$\lambda_{identity}$,$\lambda_{lip}$,$\lambda_{BL}$ are the hyperparameters used to control the importance of various loss functions in the above objective function

\section{Experiments and Results}

\subsection{Datasets \& Training}

Our model is implemented in Pytorch and takes approximately $4$ days to run on $4$ Nvidia V100 GPUs for training. Around $5000$ and $1200$ videos of the GRID dataset are used for training and testing purposes. We have taken $3000$ and $600$ videos of the LOMBARD GRID and CREMA-D datasets for training and testing purposes respectively. The frames are extracted at $25$fps. We have taken 16khz as sampling frequency for audio signals and used $13$MFCC coefficients for 0.2 sec of overlapping audio for experimentation.

We have used the GRID dataset~\citep{Alpher03}, LOMBARD GRID~\citep{gridlombard} and CREMA-D~\citep{crema} for the experimentation and evaluation of different metrics for stage 1.

We have used the GRID dataset~\citep{Alpher03} and hikemoji animated videos for experimentation and evaluation of different metrics. We have used around 100 videos for training and 30 videos for testing purposes in both the domains respectively for Stage 2.

 \subsection{Metrics}

To quantify the quality of the final generated video, we use the following metrics. PSNR(Peak Signal to Noise Ratio), SSIM(Structural Similarity Index), CPBD(Cumulative Probability Blur Detection), ACD(Average Content Distance) and KID(Kernel Inception Distance). KID~\citep{kid} computes the squared Maximum Mean Discrepancy between the feature representations of real and generated images. PSNR, SSIM, and CPBD measure the quality of the generated image in terms of the presence of noise, perceptual degradation, and blurriness respectively. ACD~\citep{Tulyakov:2018:MoCoGAN} is used for the identification of the speaker from the generated frames by using OpenPose~\citep{cao2018openpose}. Along with image quality metrics, we also calculate WER(Word Error Rate) using pretrained LipNet architecture~\citep{assael2016lipnet} and Blinks/sec using ~\citep{Authors14} to evaluate our performance of speech recognition and eye-blink reconstruction respectively.

 \subsection{Qualitative Results}
 
  OneShotAu2AV produces natural-looking high-quality animated videos of an unseen input image and audio signals. The videos are able to do lip synchronization on the sentences provided to them as well add natural expressions such as head movements, eye-blink, eyebrow movements. Videos were generated targeting different languages ensuring the proposed method is language independent and can generate videos for any linguistic community. \\

\begin{figure}[h!]%
    \centering
    \subfloat{{\includegraphics[width=6cm]{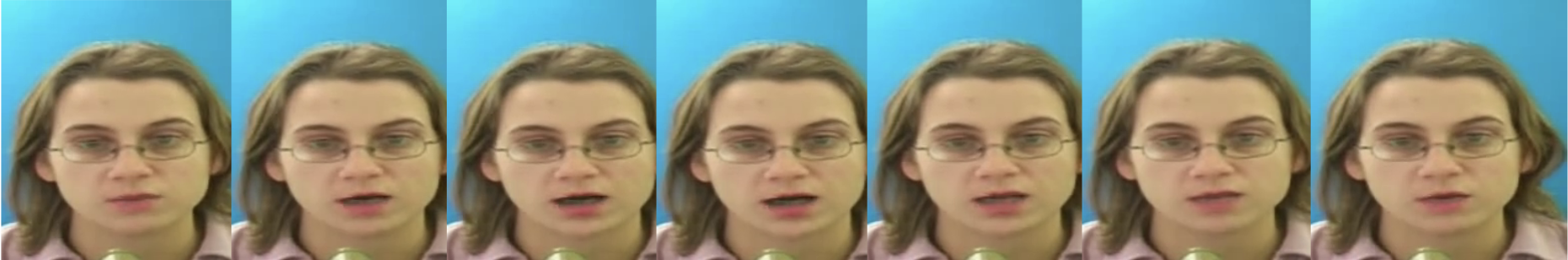} }}%
    \qquad
    \subfloat{{\includegraphics[width=6cm]{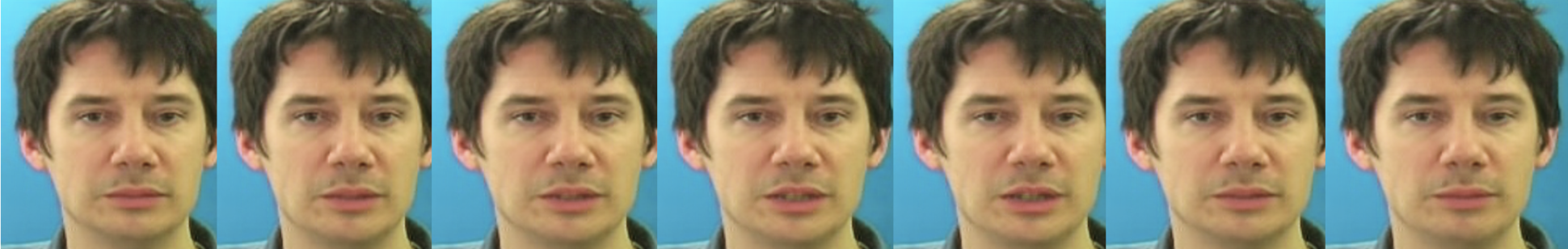} }}%
    \caption{\abovecaptionskip \belowcaptionskip Left side: Female uttering the word "now"; Right side: Male uttering the word "bin"}%
    \label{fig:bin-now}%
\end{figure}

\begin{figure}[h!]%
    \centering
    \subfloat{{\includegraphics[width=6cm]{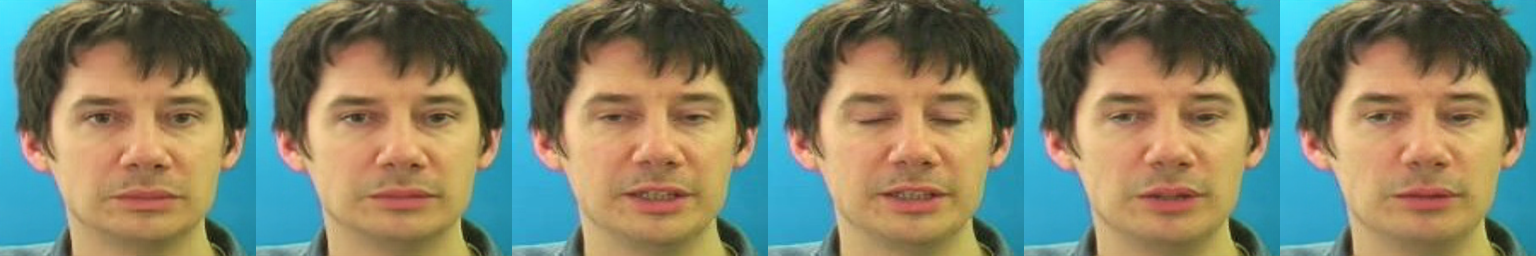} }}%
    \qquad
    \subfloat{{\includegraphics[width=6cm]{images/eye_blink.png} }}%
    \caption{ Left side: Blinking of eyes of the person while speaking; Right side: Man with facial frowns.}%
    \label{fig:blink}%
\end{figure}

\begin{figure}[h!]%
    \centering
    \subfloat{{\includegraphics[width=6cm]{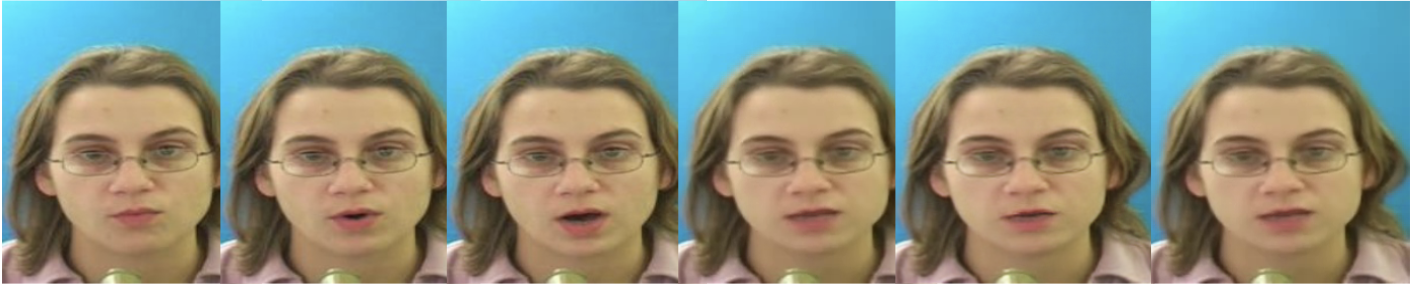} }}%
    \qquad
    \subfloat{{\includegraphics[width=6cm]{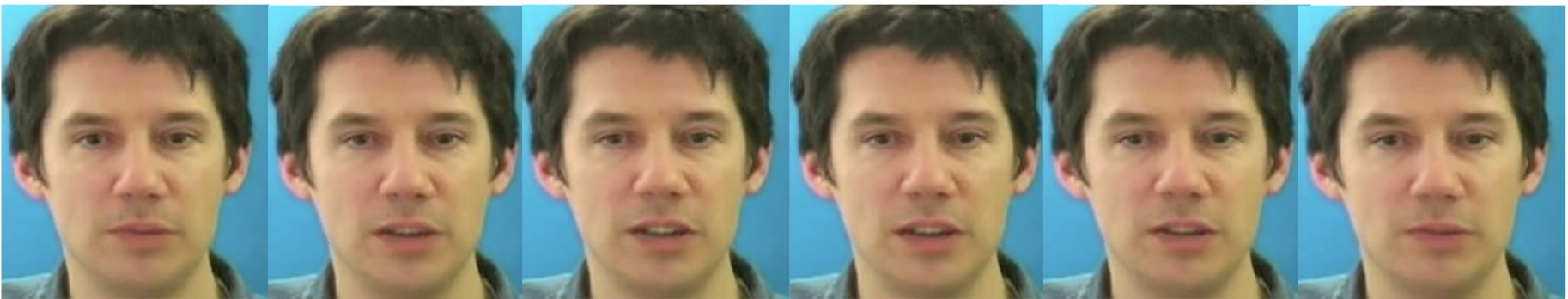} }}%
    \caption{Left side: Generated output for a Hindi audio clip ("Modi"); Right side: Generated output for a Bengali audio clip ("aache").}%
    \label{fig:words}%
\end{figure}

\begin{figure}[h!]%
    \centering
    \subfloat{{\includegraphics[width=6cm]{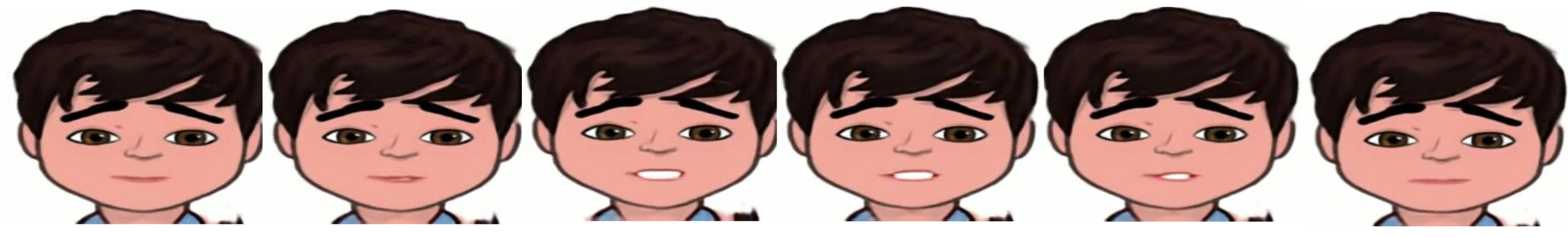} }}%
    \qquad
    \subfloat{{\includegraphics[width=6cm]{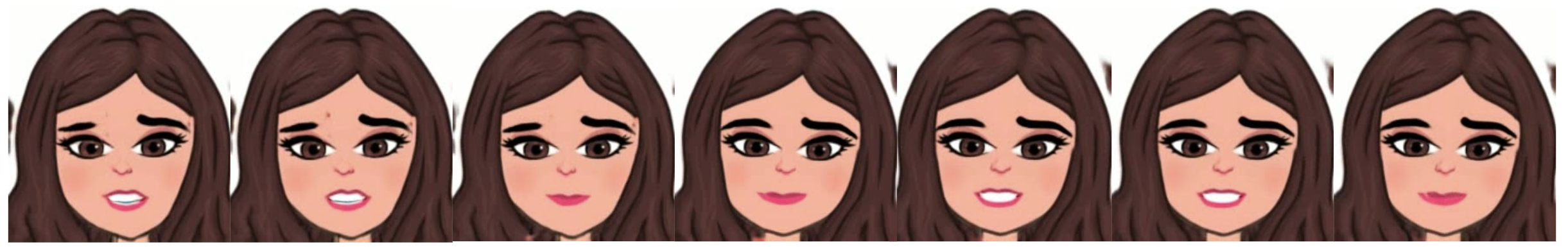} }}%
    \caption{Left side: Animated output speaking 'now'; Right side: Head movement of female anime.}%
    \label{fig:anime1}%
\end{figure}

\begin{figure}[h!]%
    \centering
    \subfloat{{\includegraphics[width=6cm]{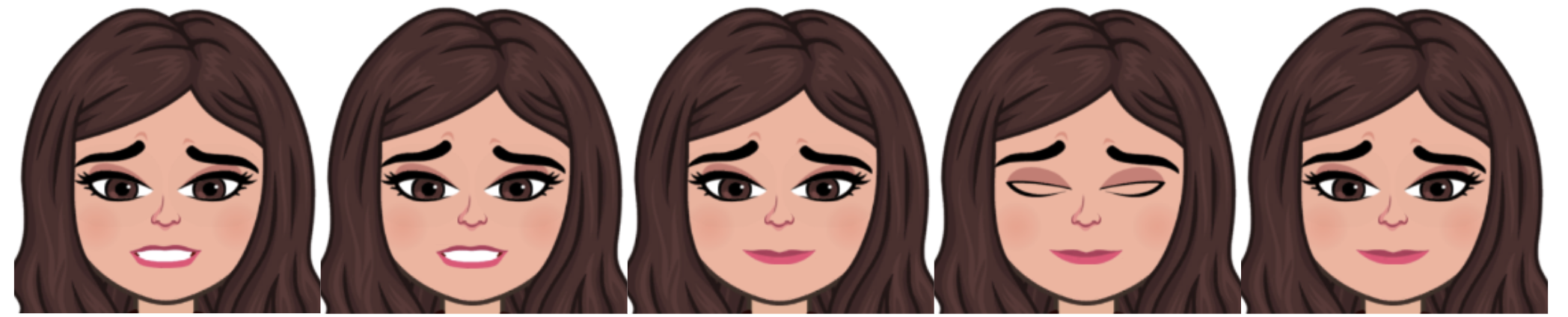} }}%
    \qquad
    \subfloat{{\includegraphics[width=6cm]{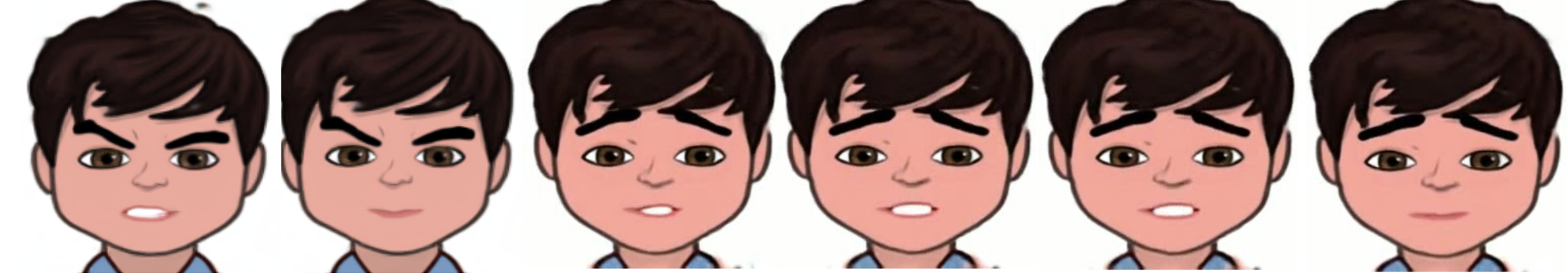} }}%
    \caption{Left side: Animated output with eye blinks; Right side:  Eyebrow movements of male while speaking.}%
    \label{fig:anime2}%
\end{figure}

\begin{figure}[h!]%
    \centering
    \subfloat{{\includegraphics[width=6cm]{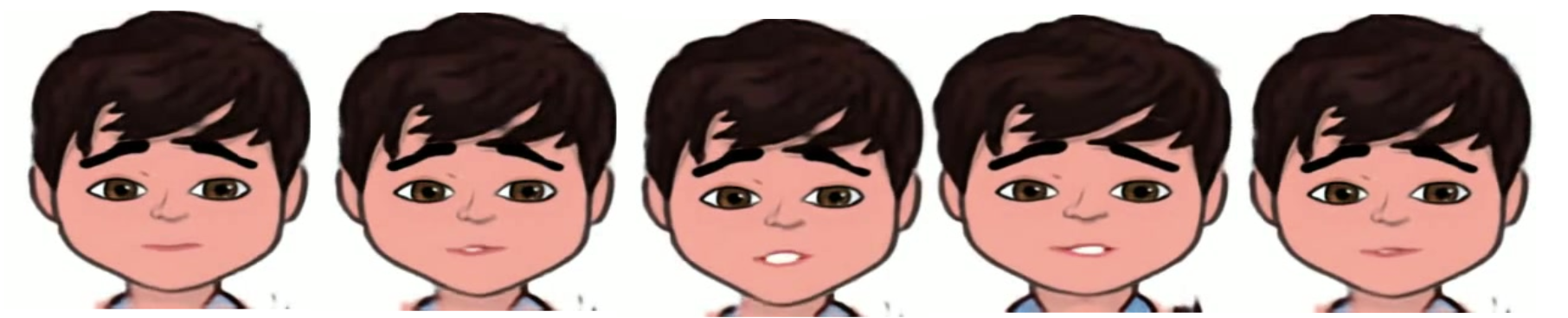} }}%
    \qquad
    \subfloat{{\includegraphics[width=6cm]{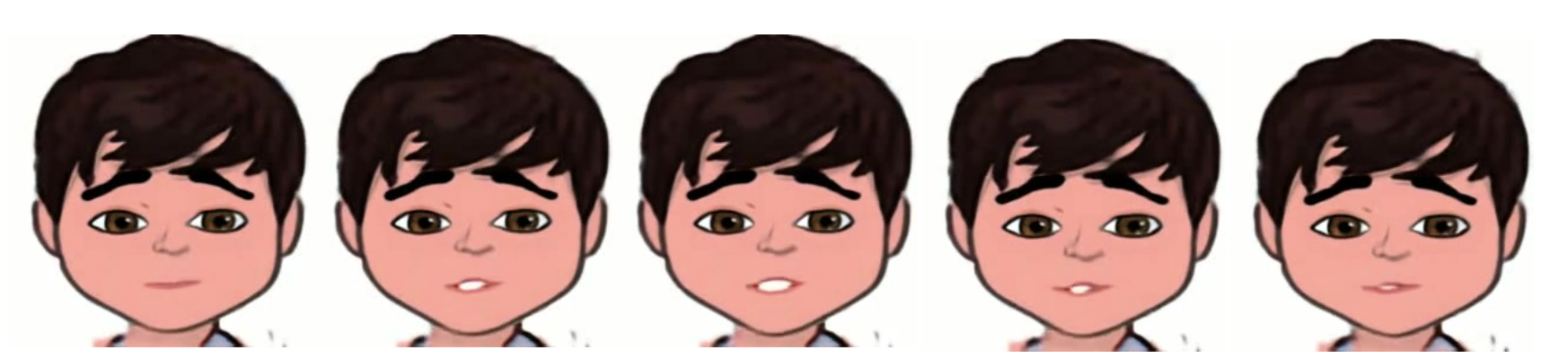} }}%
    \caption{Left side: Anime speaking the Hindi word 'modi'; Right side: Anime speaking the Bengali word 'aache'.}%
    \label{fig:anime3}%
\end{figure}

 Figure ~\ref{fig:bin-now} and Figure ~\ref{fig:anime1} display different examples of generated lip-synchronized video for male and female test cases for human and animated domains. As observed the opening and closing of the mouth is in sync with the audio signal. Figure ~\ref{fig:anime1} also displays a slight head movement of the animated person(between frames 3 and 4). Figure ~\ref{fig:blink} and  Figure ~\ref{fig:anime2} display eye-blink and facial expressions such as frowns in the both domain videos. Figure ~\ref{fig:words} displays different examples of generated lip-synchronized video for people uttering words of Hindi and Bengali language words, such as 'Modi' and 'aache' respectively. Figure ~\ref{fig:anime3} displays the same for animated domain.

  \begin{table}[h!]
  \begin{center}
  \graphicspath{ {./images/} }
    \begin{tabular}{c|c} 
      \textbf{Method} & \textbf{Results speaking 'now'}\\
      \hline
       Input frames&  \includegraphics[width=6cm]{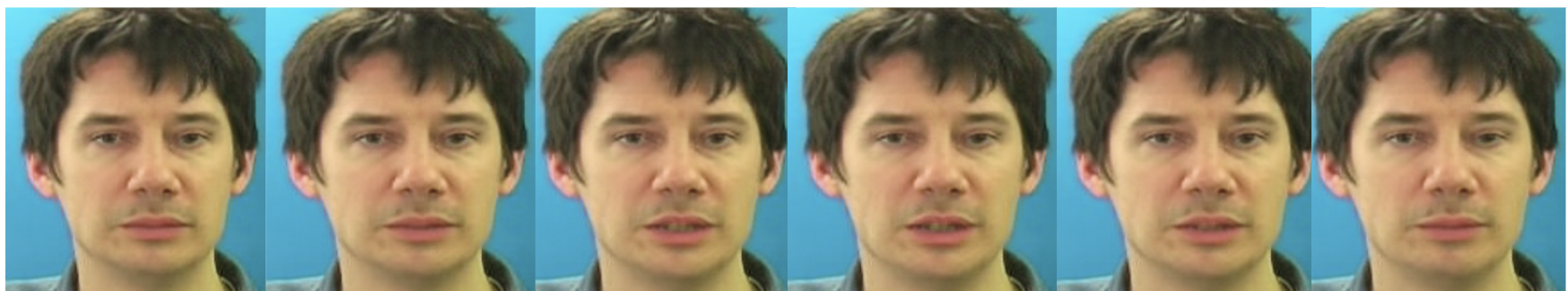} \\
       OneShotAu2AV & \includegraphics[width=6cm]{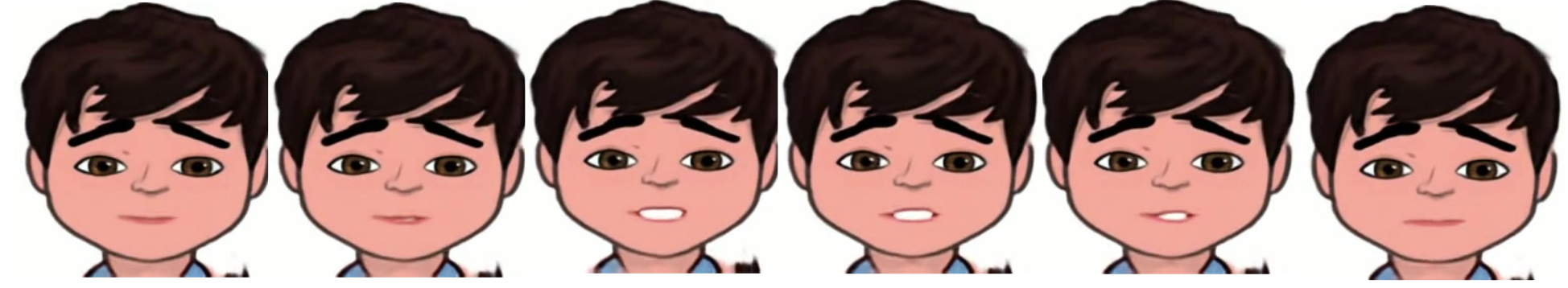}  \\
       U-GAT-IT  & \includegraphics[width=6cm]{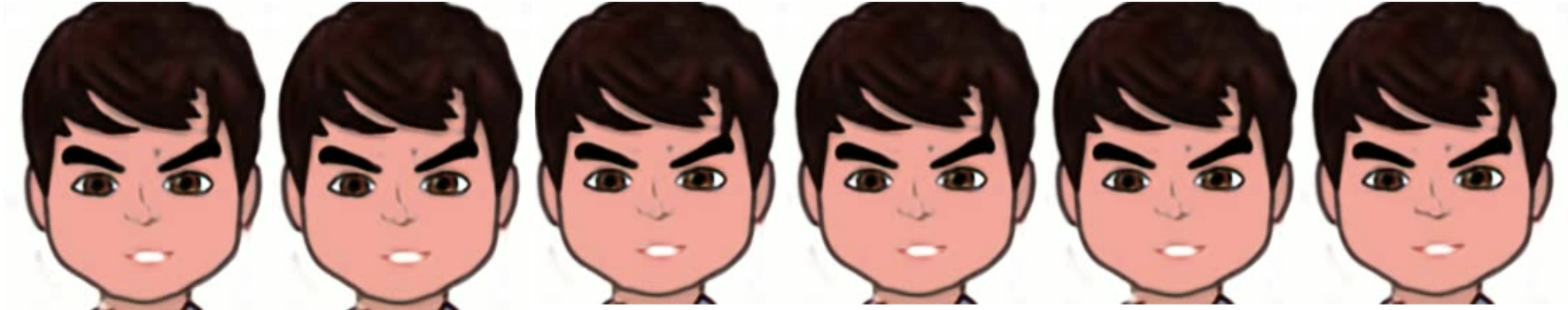} \\
       RecycleGAN    & \includegraphics[width=6cm]{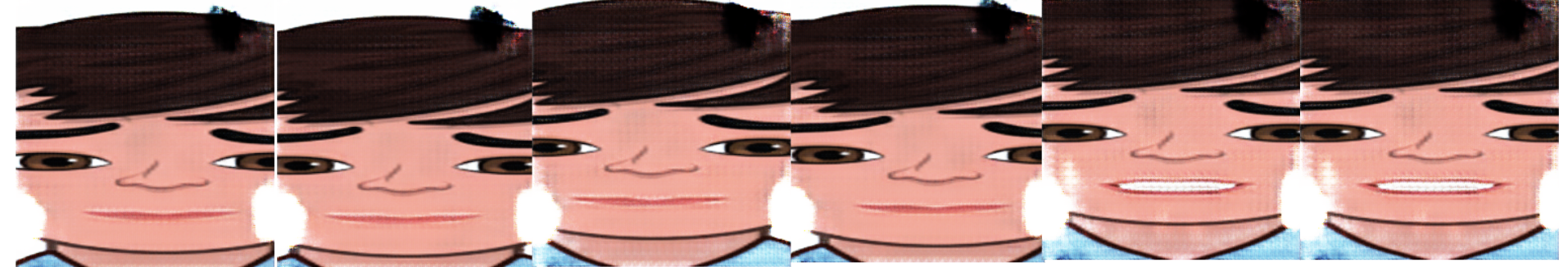}  \\
    \end{tabular}
    \vspace {0.25\baselineskip}
    \caption{Comparision of OneShotAu2AV with U-GAT-IT and RecycleGAN}
    \label{tab:tablecom}
  \end{center}
\end{table}

  \subsection{Quantitative Results}

 \paragraph{Stage 1}The proposed model performs better on image reconstruction metrics including PSNR and SSIM for both GRID and CREMA-D datasets as compared to Realistic Speech-Driven Facial Animation with GANs(RSDGAN)~\citep{Alpher05} and Speech2Vid~\citep{Chung17b} as shown in Table ~\ref{tab:table2}. The table also displays the performance of OneShotAu2AV trained on the LOMBARD GRID dataset ~\citep{gridlombard}. The improved performance of the proposed method is achieved with the use of spatially adaptive normalization in the generator architecture along with training of the proposed model in curriculum learning fashion with appropriate adversarial and non-adversarial losses.

 \begin{table}[h!]
  \begin{center}

    \begin{tabular}{c|c|c|c|c|c|c} 
      \textbf{Method} & \textbf{SSIM} & \textbf{PSNR} & \textbf{CPBD} & \textbf{WER} & \textbf{ACD-C} & \textbf{ACD-E}\\
      \hline
       OneShotAu2AV(GRID)& \textbf{0.881} & \textbf{28.571} & 0.262 &27.5 & 0.005 & 0.09  \\
       RSDGAN(GRID)  & 0.818 & 27.100 & \textbf{0.268} & \textbf{23.1} & - & 1.47x10\textsuperscript{-4}\\
       Speech2Vid(GRID)    & 0.720 & 22.662 & 0.255 & 58.2 & 0.007 & 1.48x10\textsuperscript{-4} \\
       \hline
       OneShotAu2AV(CREMA-D)& \textbf{0.773} & \textbf{24.057} & 0.184 &NA& 0.006 & 0.96  \\
       RSDGAN(CREMA-D)  & 0.700 & 23.565 & 0.216 & NA & - & 1.40x10\textsuperscript{-4}\\
       Speech2Vid(CREMA-D)    & 0.700 & 22.190 & \textbf{0.217} & NA & 0.008 & 1.73x10\textsuperscript{-4}\\
       \hline
      OneShotAu2AV(lombard) & \textbf{0.922} & \textbf{28.978} & \textbf{0.453} & \textbf{26.1} &\textbf{0.002} &\textbf{0.064}\\
    Speech2Vid(lombard) & 0.782 & 26.784 & 0.406 & 53.1 &0.004 &0.069\\
    \end{tabular}
    \vspace {0.25\baselineskip}
    \caption{Comparision of OneShotAu2AV with RSDGAN and Speech2Vid for GIRD, GRID lombard and CREMA-D datasets for SSIM, PSNR, CPBD, WER and ACD by calculating cosine distance(ACD-C)(should be 0.02 and below) and euclidean distance(ACD-E)(should be 0.2 and below).}
    \label{tab:table2}
    
  \end{center}
\end{table}

\paragraph{Stage 2:}The proposed Model shows better results in terms of image translation metric i.e KID is 2x and 8x better than U-GAT-IT and RecycleGan respectively as displayed in Table ~\ref{tab:table4}. This is achieved by adding temporal predictor based recycle loss and lip-sync loss in the networks which helps in the reconstruction of high quality animated output. OneShotAu2AV performs better on lip synchronization metric(WER) which is 2x and 8x better than U-GAT-IT and RecycleGan respectively.

 \begin{table}[h!]
  \begin{center}

    \begin{tabular}{c|c|c|c} 
      \textbf{Method} & \textbf{KID$\times$100$\pm$std.$\times$100}  & \textbf{WER} & \textbf{Blink/sec}\\
      \hline
       OneShotAu2AV& \textbf{5.02$\pm$0.03}  & \textbf{31.97} &\textbf{0.546}  \\
       U-GAT-IT  & 10.37$\pm$0.17 & 68.85 & 0.046  \\
       RecycleGan & 42.54$\pm$0.68 & 240.51 & 0.0   \\
    \end{tabular}
    \vspace {0.25\baselineskip}
    \caption{Comparision of OneShotAu2AV with U-GAT-IT and RecycleGAN for KID, WER and  Blink/sec.}
    \label{tab:table4}
    
  \end{center}
\end{table}

\subsection{Ablation Study}

We conducted detailed ablation studies.In Stage 1, the addition of Contrastive Loss and multi-scale temporal adversarial loss leads to the improvement of the SSIM(0.867 to 0.873), PSNR(27.996 to 28.327) and CPBD(0.213 to 0.259) scores when measured on GRID dataset. Adding blink loss leads to further improvement in SSIM(0.873 to 0.881), PSNR(28.327 to 28.571), CPBD(0.259 to 0.262) scores. Similar improvements are observed on the LOMBARD GRID dataset as well.
 
In Stage 2, the addition of predictive model based recycle loss and lip-sync loss helped in the improvement of KID (10.37 to 5.02) and WER (68.85 to 31.97). For further details, kindly refer to the supplementary material.

\paragraph{Psychophysical assessment:}For video attachments and Psychophysical assessment(results of Turing test and user ratings), kindly refer to the supplementary material.

\section{Conclusion and Future Work}
In this paper, we have presented a novel approach, \textit{OneShotAu2AV}, to convert an audio and single image of a person to an animated video. Using two stages in our multi-level generators and discriminators based architecture and appropriate adversarial and non-adversarial losses, we are able to achieve synced lip movements, blinks, and eye-brow movements in the output. Experimental evaluation demonstrates superior performance of OneShotAu2AV as compared to U-GAT-IT and RecycleGan on multiple quantitative metrics including KID(Kernel Inception Distance), Word error rate, blinks/sec. In future, we will look at techniques to further enhance the expressiveness of the generated animated videos.


\end{document}


\maketitle
\section{Architectural Design}

\subsection{Stage 1}

This Stage consists of converting and audio and an unseen person$'$s image to a human video. 

\subsubsection{Audio Pre processing}
 An audio input of 200 ms is given along with the image to produce a single frame of the video. The audio input is overlapping with the previous audio input with an overlapping interval of 0.16 ms. Every audio frame is centered around a single video frame. To do that, zero padding is done before and after the audio signal and uses the following formula for the stride.
 \begin{align*}
      stride = \frac{\text{audio sampling rate}}{\text{video frames per sec}}
\end{align*}

\begin{figure}[h!]
  \begin{center}
   \includegraphics[width=0.6\linewidth]{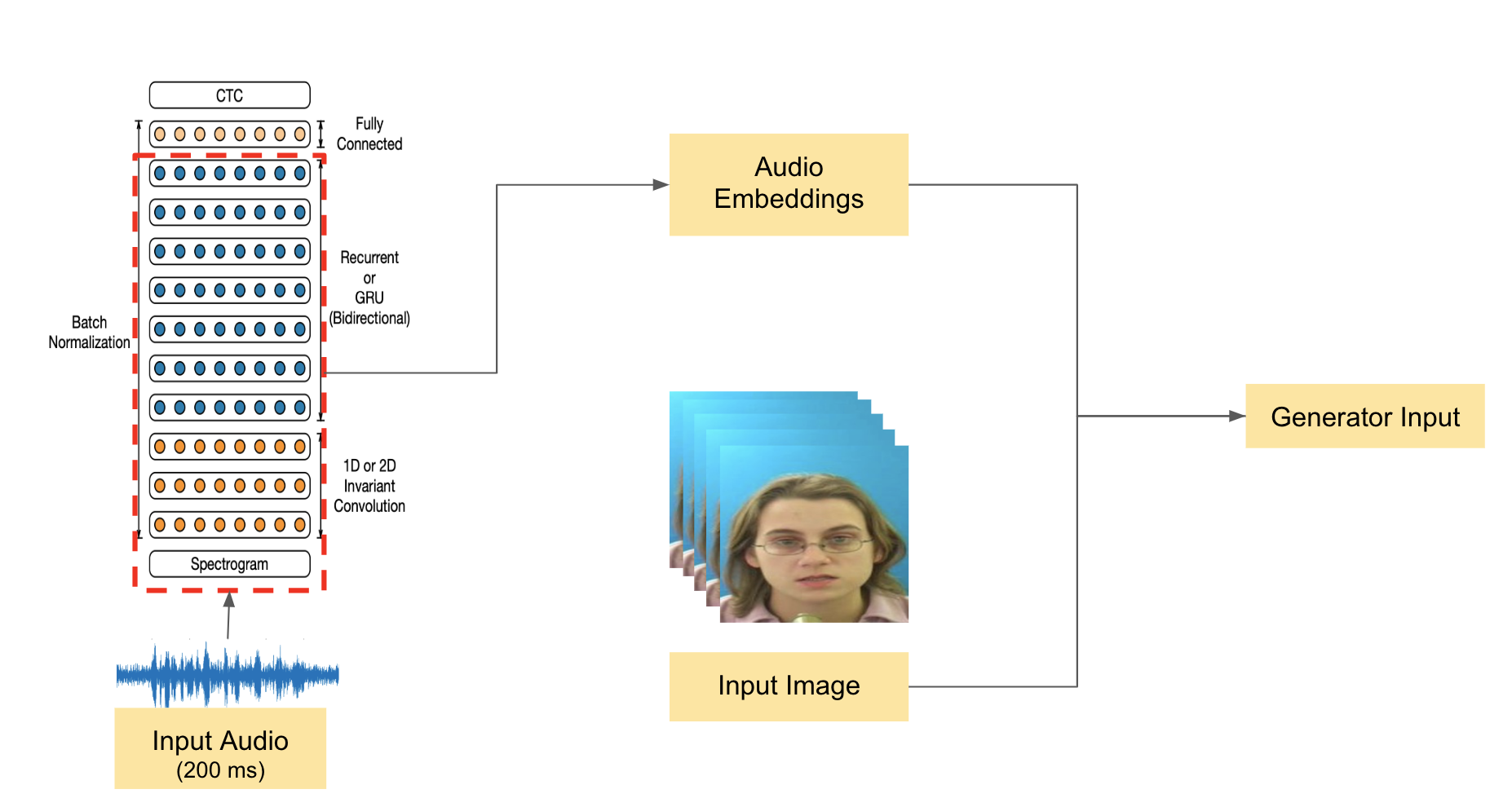}
   \caption{Audio signals are fed into deep speech2 architecture to extract content embeddings }
   \label{fig:deep}
  \end{center}
\end{figure}

The MFCC value of the audio segment is fed into a deepspeech2 model to extract the content related features which then goes to the generator at stage 1. The figure is mention in ~\ref{fig:deep}

\subsubsection{Synchronization Discriminator}

The architecture of the discriminator is given in figure ~\ref{fig:sync}(a) which contains the input an audio signal of 200ms time interval(5 audio signals of 40ms each) and 5 frames of the video. The lower half of the frame of resized to (224,224,3) is fed as an input. This loss is fed back to the generator of our model to learn coherent lip synchronization.

\subsubsection{ Multi-scale discriminator~\citep{wang2018pix2pixHD}}

Multi-scale discriminator is used in multi scale frame discriminator and multi scale temporal discriminator which consists of 3 discriminators that have an identical network structure but operate at different image scales. These discriminators are referred to as D1, D2, and D3. Specifically, we downsample the real and synthesized high-resolution images by a factor of 2 and 4 to create an image pyramid of 3 scales. The discriminators D1, D2, and D3 are then trained to differentiate real and synthesized images at the 3 different scales, respectively. The discriminators operate from coarse to fine level and help the generator to produce high-quality images.

\subsubsection{Losses}
OneShotAu2AV is trained with different losses to generate realistic videos as explained below.

\begin{figure}%
    \centering
    \subfloat{{\includegraphics[width=5cm]{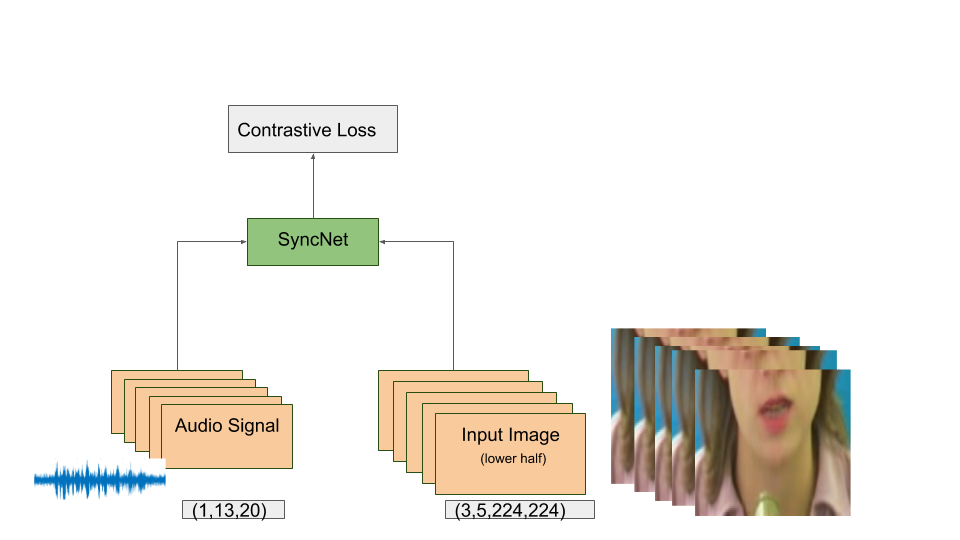} }}%
    \qquad
    \subfloat{{\includegraphics[width=5cm]{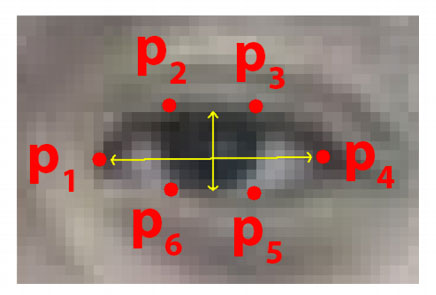} }}%
    \caption{(a) SyncNet architecture for better lip synchronization which is trained on GRID dataset with contrastive loss and then used its loss in our proposed architecture. (b)Description of the 6 eye points. p represents the landmark points of eyes}.%
    \label{fig:sync}%
\end{figure}

\paragraph{Adversarial Loss}
Adversarial Loss is used to train the model to handle adversarial attacks and ensure the generation of high-quality images for the video. The loss is defined as:

\begin{align*}
    L\textsubscript{GAN}(G,D) &= E\textsubscript{x$\sim$P\textsubscript{d}}[\log(D(x))] + E\textsubscript{z$\sim$P\textsubscript{z}}[\log(D(1-G(z)))] 
\end{align*}

 where G tries to minimize this objective against an adversarial D that tries to maximize where z is the distribution of source(image) and x is the distribution of the video frames.

\paragraph{Reconstruction loss}
Reconstruction loss~\citep{RLoss} is used on the lower half of the image to improve the reconstruction in the mouth area. L1 loss is used for this purpose as described below:

\begin{align*}
    L\textsubscript{RL} &= \sum_{n\epsilon [0,W]*[H/2,H]}^{}(R\textsubscript{n} - G\textsubscript{n})
\end{align*}

where, R\textsubscript{n} and G\textsubscript{n} are the real and generated frames respectively and W and H represent the height and width of an image.

\paragraph{Feature Loss}

Feature-matching Loss~\citep{wang2018pix2pixHD} ensures generation of natural-looking  high-quality frames. We take the L1 loss of between generated images and real images for different scale discriminators and then sum it all. We extract features from multiple layers of the discriminator and learn to match these intermediate representations from the real and the synthesized image. This helps in stabilizing the training of the generator. The feature matching loss,   L\textsubscript{FM}(G,D\textsubscript{k}) is given by:

\begin{align*}
    L\textsubscript{FM}(G,D\textsubscript{k}) &= E\textsubscript{(x,z)} \sum_ {n=1}^{T}[\frac{1}{N\textsubscript{i}}||D\textsubscript{k}^{(i)}(x)-D\textsubscript{k}^{(i)}(G(z))||\textsubscript{1}]
\end{align*}

where, T is the total number of layers and N\textsubscript{i} denotes the
the number of elements in each layer.

\paragraph{Perceptual Loss}
The perceptual similarity metric is calculated between the generated frame and the real frame. This is done by using features of a VGG19~\citep{VGG19} model trained for ILSVRC classification and VGGFace~\citep{VGGFace} dataset.The perceptual loss~\citep{PerceptualLoss},(L\textsubscript{PL}) is defined as:

\begin{align*}
    L\textsubscript{PL} &= \lambda\sum_{n=1}^{N}[\frac{1}{M\textsubscript{i}}||F^{(i)}(x)-F^{(i)}(G(z))||\textsubscript{1}]
\end{align*}
where, $\lambda$ is the weight for perceptual loss and $F^{(i)}$ is the ith layer of VGG19 network with M\textsubscript{i} elements of VGG layer.

\paragraph{Contrastive Loss}
For coherent lip synchronization, we use the Synchronization Discriminator with Contrastive loss. The training objective is that the output of the audio and the video networks are similar for genuine pairs, and different for false pairs.

Contrastive loss,(L\textsubscript{CL}) is given by following equation 
\begin{align*}
    L\textsubscript{CL} &= \frac{1}{2N}\sum_{n=1}^{N}(y\textsubscript{n})d^2\textsubscript{n}+(1-y\textsubscript{n})max(margin-d\textsubscript{n},0)^2 &
\end{align*}

\begin{align*}
    d\textsubscript{n} &= ||v\textsubscript{n}-a\textsubscript{n}||\textsubscript{2}
\end{align*}

where, v\textsubscript{n} and a\textsubscript{n}  are fc\textsubscript{7} vectors for video and audio inputs respectively. y $\epsilon$ [0,1] is the binary similarity metric for video and audio input.

\paragraph{Blink loss }

We have used the eye aspect ratio (EAR) taken from Real-Time Eye Blink Detection using Facial Landmarks ~\citep{Authors14} to calculate the blink loss. A blink is detected at the location where a sharp drop occurs in the EAR signal. Loss is defined as:

\begin{align*}
  m &= \frac{||p2-p6|| + ||p3-p5||}{||p1-p4||}
\end{align*}

\begin{align*}
    L\textsubscript{BL} &= ||m\textsubscript{r}-m\textsubscript{g}||
\end{align*}

where, p\textsubscript{i} is described in Figure ~\ref{fig:sync}(b). We have taken the L1 loss of eye aspect ratio(EAR) between real image m\textsubscript{r}  and synthesized frame m\textsubscript{g}.

\subsection{Stage 2}
This stage consists of converting a human video into an animated video.

\subsubsection{ Temporal Predictor}
\begin{figure}[h!]
  \begin{center}
   \includegraphics[width=0.6\linewidth]{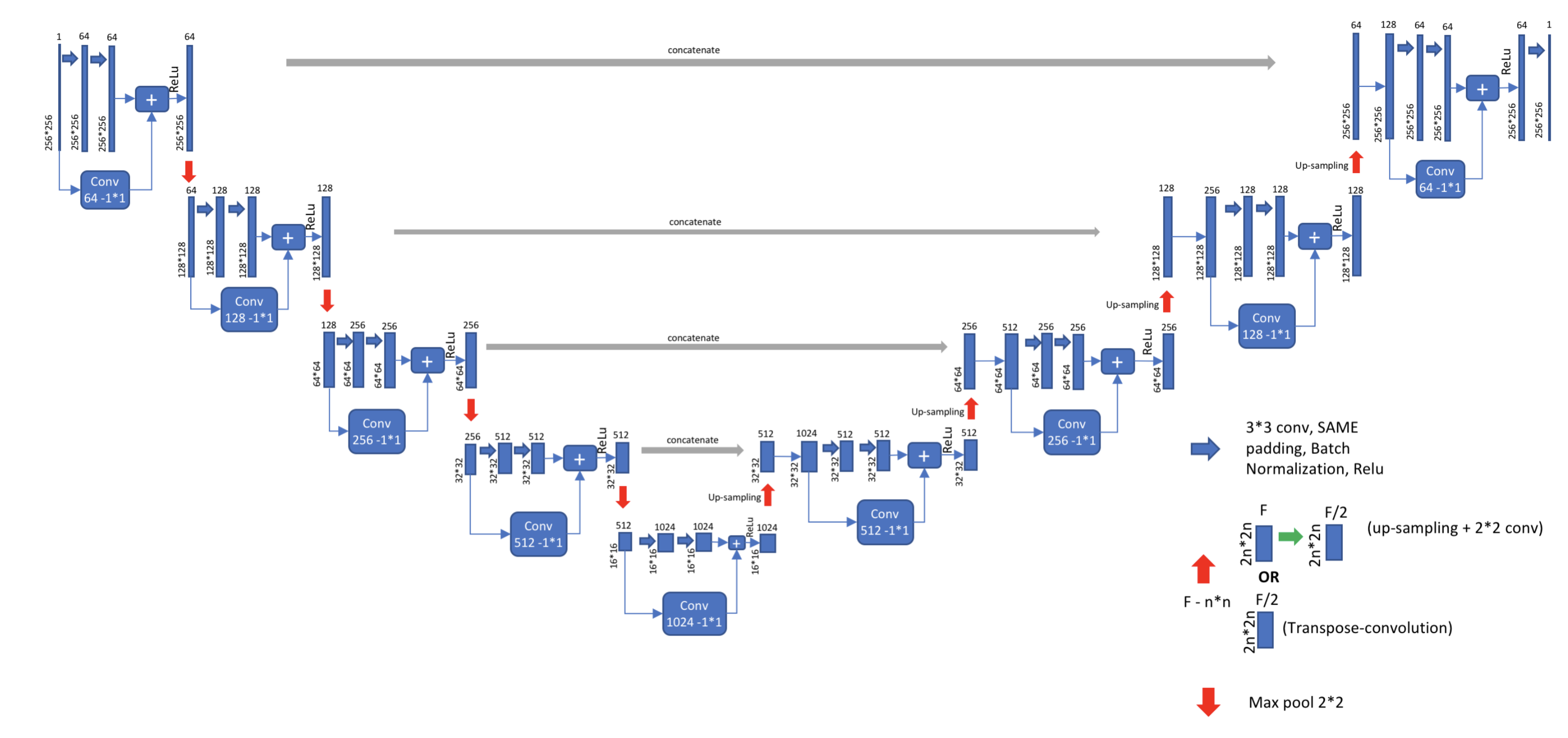}
   \caption{Unet architecture}
   \label{fig:unet}
  \end{center}
\end{figure}

The temporal predictor is used in the second stage is to capture the temporal information from the video which is trained with L2 loss and the equation is given below. The temporal predictor has the UNet architecture which is given in figure ~\ref{fig:unet}

\begin{align*}
 L\textsubscript{P\textsubscript{x}} &= \sum_{t}^{}(||x\textsubscript{t+1} - P\textsubscript{x}(x\textsubscript{1:t})||^2)
 \end{align*}
 which is given below where x\textsubscript{1:t} is (x\textsubscript{1},x\textsubscript{2},....x\textsubscript{t}).

\subsubsection{Losses}

Different loss functions are used to create the animated videos capturing various aspects such as lip-syncing, eye blinking, eyebrows movements, and head movements.

\paragraph{Adversarial Losses:}
This is used to improve the mapping of a translated image to a target image. 
\begin{align*}
    L\textsubscript{GAN}(G,D) &= E\textsubscript{y$\sim$P\textsubscript{t}}[\log(D(y))^2] + E\textsubscript{x$\sim$P\textsubscript{s}}[\log(D(1-G(x)))^2] 
\end{align*}

 where G tries to minimize this objective against an adversarial D that tries to maximize and x and z belongs to source and target domain respectively.
 
 \paragraph{Recylce loss:}
We use the temporal predictor,P\textsubscript{x} to define the recycle loss to preserve the temporal coherency across the generated animated videos and to avoid the perceptual mode collapse i.e. for different frames of real videos as an input, we are getting same animated output frames. 
\begin{align*}
  L\textsubscript{recycle}(G\textsubscript{x},G\textsubscript{y},P\textsubscript{y}) &= \sum_{t}^{}(||x\textsubscript{t+1} - G\textsubscript{x}(P\textsubscript{y}(G\textsubscript{y}(x\textsubscript{1:t})))||^2)
  \end{align*}
    where x and y are the source and target domain respectively and $G\textsubscript{y}(x\textsubscript{1:t})$ is $(G\textsubscript{y}(x\textsubscript{1})$,G\textsubscript{y}(x\textsubscript{1}),...,G\textsubscript{y}(x\textsubscript{t})), Intuitively, this will learn the sequence to frames to map back to themselves.

\paragraph{Identity loss:}
This is used to ensure the color distribution of the source and target image to be similar.

 \begin{align*}
  L\textsubscript{identity} &= E\textsubscript{x$\sim$P\textsubscript{s}}(| x -  (G\textsubscript{s->t}(x))|\textsubscript{1})]
  \end{align*}

\paragraph{CAM loss:}
By exploiting the information from the auxiliary classifiers , n\textsubscript{s} and n\textsubscript{D\textsubscript{t}}, G\textsubscript{s->t} and D\textsubscript{t}  need to know where they need to improve and what  makes the most difference between two domain.
 \begin{align*}
    L\textsubscript{cam}^{s->t} &= E\textsubscript{y$\sim$P\textsubscript{t}}[\log(n\textsubscript{s}(y))] + E\textsubscript{x$\sim$P\textsubscript{s}}[\log(D(1-n\textsubscript{s}(x)))] 
\end{align*}
 \begin{align*}
    L\textsubscript{cam}^{D\textsubscript{t}} &= E\textsubscript{y$\sim$P\textsubscript{t}}[\log(n\textsubscript{D\textsubscript{t}}(y))] + E\textsubscript{x$\sim$P\textsubscript{s}}[\log(D(1-n\textsubscript{D\textsubscript{t}}(G\textsubscript{s->t}(x))))] 
\end{align*}

 \paragraph{Lip sync Loss:}
We have used the cycle consistency loss on the lower half between x and G\textsubscript{t->s}(G\textsubscript{s->t}(x)) as proposed 
 \begin{align*}
  L\textsubscript{lip} &= \sum_{n\epsilon [0,W]*[H/2,H]}^{}(| x -  G\textsubscript{t->s}(G\textsubscript{s->t}(x))|\textsubscript{1})]
  \end{align*}
 \paragraph{blink Loss:}
We have used the eye aspect ratio (EAR) taken from Real-Time Eye Blink Detection using Facial Landmarks ~\citep{Authors14} to calculate the blink loss.Loss is defined as 
\begin{align*}
  m &= \frac{||p2-p6|| + ||p3-p5||}{||p1-p4||}
\end{align*}

\begin{align*}
    L\textsubscript{BL} &= ||m\textsubscript{r}-m\textsubscript{g}||
\end{align*}
where, p\textsubscript{i} is described in Figure ~\ref{fig:sync}(b). We have taken the L1 loss of eye aspect ratio(EAR) between real image m\textsubscript{x}  and m\textsubscript{G\textsubscript{t->s}(G\textsubscript{s->t}(x))}.

\section{Experiments}

\subsection{Metrics}

\paragraph{1. PSNR- Peak Signal to Noise Ratio:} It computes the peak signal to noise ratio between two images. The higher the PSNR the better the quality of the reconstructed image.

\paragraph{2. SSIM- Structural Similarity Index:} It is a perceptual metric that quantifies image quality degradation.\ The larger the value the better the quality of the reconstructed image.

\paragraph{3. CPBD- Cumulative Probability Blur Detection:} It is a perceptual based no-reference objective image sharpness metric based on the cumulative probability of blur detection developed at the Image.

\paragraph{4. WER- Word error rate:} It is a metric to evaluate the performance of speech recognition in a given video. We have used LipNet architecture~\citep{assael2016lipnet} which is pre-trained on the GRID dataset for evaluating the WER. On the GRID dataset, Lipnet achieves 95.2 percent accuracy which surpasses the experienced human lipreaders.

\paragraph{5. ACD- Average Content Distance(~\citep{Tulyakov:2018:MoCoGAN}):} It is used for the identification of speakers from the generated frames using OpenPose~\citep{cao2018openpose}. We have calculated the Cosine distance and Euclidean distance of representation of the generated image and the actual image from Openpose. The distance threshold for the OpenPose model should be 0.02 for Cosine distance and 0.20 for Euclidean distance ~\citep{acd2}. The lesser the distances the more similar the generated and actual images.

\paragraph{6. KID - Kernel Inception Distance~\citep{kid}:} It computes the squared Maximum Mean Discrepancy between the feature representations of real and generated images. In contrast to the Frchet Inception Distance~\citep{fid}, KID has an unbiased estimator, which makes it more reliable, especially when there are fewer test images than the dimensionality of the inception features. The lower KID indicates that the more shared visual similarities between real
and generated images~\citep{kidlower}.

\paragraph{7. Blinks/sec:} To capture the blinks in the video, we are calculating the blinks/sec so that we can better understand the quality of animated videos. We have used SVM and eye landmarks along with Eye aspect ratio used in  Real-Time Eye Blink Detection using Facial Landmarks ~\citep{Authors14} to detect the blinks in a video.

\subsection{Detailed Description of Datasets}

GRID dataset is a large multi-talker audiovisual sentence corpus. This corpus consists of high-quality audio and video $($facial$)$ recordings of 1000 sentences spoken by each of 34 talkers $($18 male, 16 female$)$. LOMBARD GRID dataset is a bi-view audiovisual Lombard speech corpus that can be used to support joint computational-behavioral studies in speech perception. The corpus includes 54 talkers, with 100 utterances per talker $($50 Lombard and 50 plain utterances$)$. It consists of 5400 videos generated on  54 talkers comprising 30 female talkers and 24 male talkers. CREMA-D  is a data set of 7,442 original clips from 91 actors with six different emotions (Anger, Disgust, Fear, Happy, Neutral, and Sad).The hikemoji dataset is used for the style transfer and creating the animated videos.

\subsection{Detailed Training }

\subsubsection{Stage 1}

 The aligned face is generated for every speaker using facial landmark detector~\citep{inproceedings} and HopeNet~\citep{Ruiz_2018_CVPR_Workshops} for calculating the yaw, pitch and roll angles to get the most aligned faces for every speaker as an input.

 We take the Adam optimizer~\citep{Adam} with learning rate = 0.002 and $\beta_1$= 0.0 and $\beta_2$ = 0.90 for the generator and discriminators. The learning rate of the generator and discriminator is constant for 50 epochs and after that it decays to zeros in the next 100 epochs.
 
  \subsubsection{Stage 2}
 
We have used $\lambda\textsubscript{cam} = 2000, \lambda\textsubscript{recycle}= 100,\lambda\textsubscript{identity}=10,\lambda\textsubscript{lip}=100,\lambda\textsubscript{BL} = 100$ are the hyperparameters used to control the importance of various loss functions in the above objective function. Adam optimizer~\citep{Adam} with learning rate = 0.0001 and $\beta_1$= 0.5 and $\beta_2$ = 0.999 for the generator,  discriminators and temporal predictor.

\subsection{Ablation Study}

\subsubsection{Stage 1}

We studied the incremental impact of various loss functions on the LOMBARD GRID dataset and the GRID dataset. We have provided corresponding videos in supplementary data for better visual understanding. As mentioned in section 4 (Curriculum learning) each loss has a different impact on the final output video. Table ~\ref{tab:table4} and Table ~\ref{tab:table5}  depicts the impact of different losses on both datasets. The base model mentioned is the includes the adversarial gan loss, feature loss, and perceptual loss. The addition of contrastive loss and multi-scale temporal adversarial loss in sequence discriminator helps in achieving coherent lip synchronized videos and improves the SSIM, PSNR, and CPBD values. Further addition of Blink Loss, ensures improved quality of the final video. 

\begin{table}[h!]
  \begin{center}

    \begin{tabular}{c|c|c|c} 
      \textbf{Method} & \textbf{SSIM} & \textbf{PSNR} & \textbf{CPBD}\\
      \hline
       Base Model(BM)  & 0.869 & 27.996 &  0.213 \\
       BM + CL +TAL        & 0.873 & 28.327 &  0.258 \\
       BM + CL + TAL+ BL    & 0.881 & 28.571 & 0.262  \\
    \end{tabular}
    \vspace {0.25\baselineskip}
    \caption{Ablation Study on the GRID dataset where, CL is the contrastive loss ,TAL is the multi-scale temporal adversarial loss and BL is the Blink loss}
    \label{tab:table4}
    
  \end{center}
\end{table}

\begin{table}[h!]
  \begin{center}

    \begin{tabular}{c|c|c|c} 
      \textbf{Method} & \textbf{SSIM} & \textbf{PSNR} & \textbf{CPBD}\\
      \hline
       Base Model(BM)  & 0.909 & 28.656 &  0.386 \\
       BM + CL + TAL        & 0.913 & 28.712 &  0.390 \\
       BM + CL + TAL+ BL    & 0.922 & 28.978 & 0.453  \\
    \end{tabular}
    \vspace {0.25\baselineskip}
    \caption{Ablation Study on the LOMBARD GRID dataset where, CL is the contrastive loss, TAL is the multi-scale temporal adversarial loss and BL is the Blink loss}
    \label{tab:table5}
  \end{center}
\end{table}

The use of deepspeech2 to generate audio to content embeddings helped in the improvement of WER and help us reach almost similar performance as RSDGAN.

Table ~\ref{tab:table2} gives a brief about the videos attached related to the ablation study.

\begin{table}[h!]
  \begin{center}
    \begin{tabular}{c|c} 
      \textbf{Video Name} & \textbf{Model Description} \\
      \hline
Video\textunderscore11.mp4 & Output with Base model as explained in Section 6.6\\
Video\textunderscore12.mp4 & Output with Base model as explained in Section 6.6  \\
Video\textunderscore13.mp4 & Output with contrastive loss (CL) \\
Video\textunderscore14.mp4 & Output with blink loss (BL)  \\
Video\textunderscore15.mp4 & Output with blink loss (BL) \\

    \end{tabular}
    \vspace {0.5\baselineskip}
    \caption{Ablation Study Video Description}
    \label{tab:table2}
  \end{center}
\end{table}

\subsubsection{Stage 2}
We studied the incremental result of various loss functions on the GRID dataset and the hikemoji dataset. The addition of recycle loss, lip synchronization loss, and blink loss helps in the improvement of different aspects of an animated video such as lip movement, eye-blink, eyebrows, and head movements.
 
 \begin{table}[h!]
  \begin{center}

    \begin{tabular}{c|c|c|c} 
      \textbf{Method} & \textbf{KID$\times$100$\pm$std.$\times$100} & \textbf{WER} & \textbf{blinks/sec}\\
      \hline
       Base Model(BM)  & 10.37$\pm$0.17 & 68.85 &  0.046 \\
       BM + RL+ LSL         & 6.43$\pm$0.92 & 32.62 & 0.153 \\
       BM + RL + LSL+ BL    & 5.02$\pm$0.03 & 31.97 & 0.546  \\
    \end{tabular}
    \vspace {0.25\baselineskip}
    \caption{Ablation Study of Stage 2 on the GRID dataset where, Base model is U-GAT-IT architecture, RL is the recycle loss ,LSL is the lip synchronisation loss loss and BL is the Blink loss}
    \label{tab:table4}
    
  \end{center}
\end{table}

\section{Psychophysical assessment}

\subsection{Stage 1}
Results are visually rated (on a scale of 5) individually by 25 persons, on three aspects,
 lip synchronization, eye blinks and eyebrow raises and quality of video.
 The subjects were shown anonymous videos at the same time for the different audio clips for side-by-side comparison. Table ~\ref{tab:table1} clearly shows that OneShotAu2AV performs significantly better in quality and lip synchronization which is of prime importance in videos. 


\begin{table}[h!]
  \begin{center}

    \begin{tabular}{c|c|c|c} 
      \textbf{Method} & \textbf{Lip-Sync} & \textbf{Eye-blink} & \textbf{Quality}\\
      \hline
       OneShotAu2AV& 90.8 & 88.5 & \textbf{76.2}  \\
       RSDGAN  & \textbf{92.8} & \textbf{90.2} & 74.3\\
       Speech2Vid    & 90.7 & 87.7 & 72.2 \\
    \end{tabular}
    \vspace {0.25\baselineskip}
    \caption{Psychophysical Evaluation (in percentages) based on users rating}
    \label{tab:table1}
  \end{center}
\end{table}

\begin{figure}[h!]
  \includegraphics[width=\linewidth]{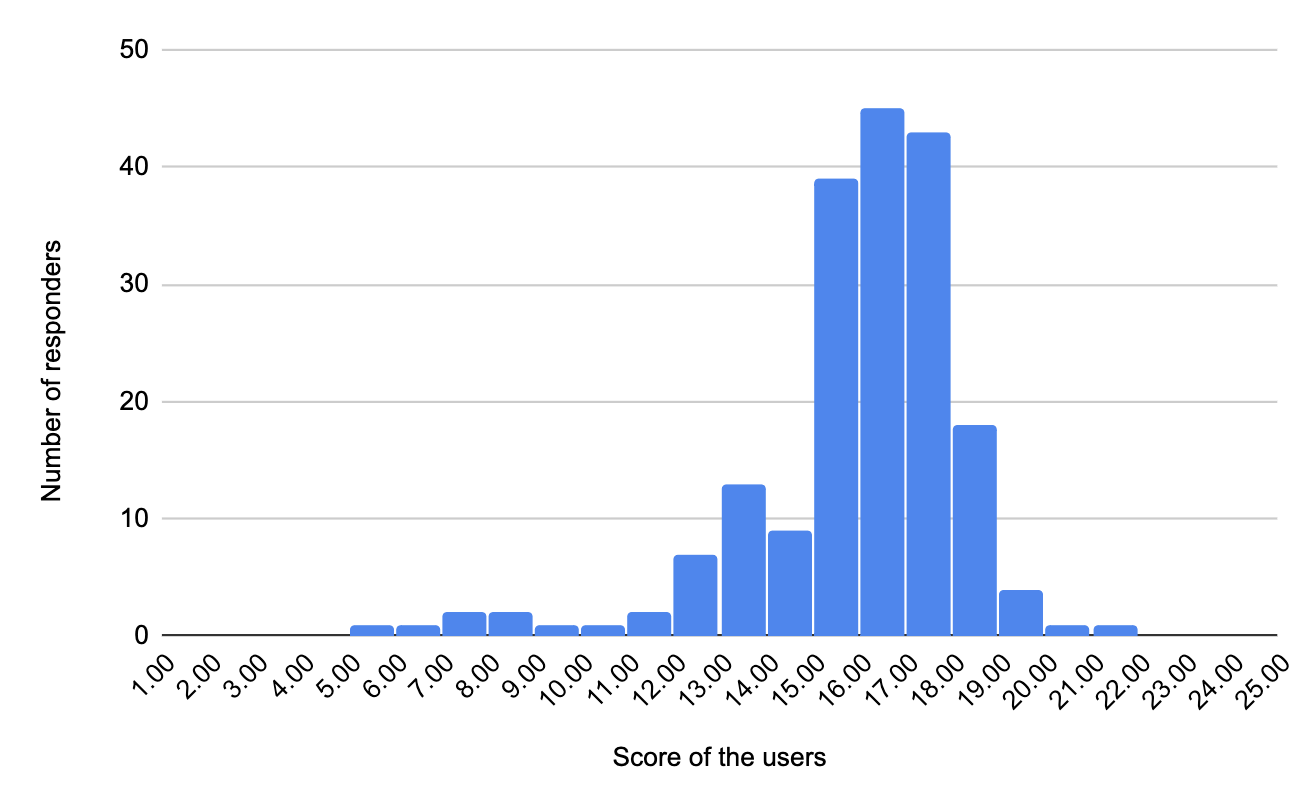}
  \caption{Distribution of user scores for the online Turing test}
  \label{fig:turing}
\end{figure}

To test the naturalism of the generated videos we conduct an online Turing test \footnote{\url{https://forms.gle/JEk1u5ahc9gny7528}}. Each test consists of 25 questions with 13 fake and 12 real videos. The user is asked to label a video real or fake based on the aesthetics and naturalism of the video. Approximately 300 user data is collected and their score of the ability to spot fake video is displayed in Figure ~\ref{fig:turing}.

\subsection{Stage 2}

Table ~\ref{tab:tableanime} clearly shows that OneShotAu2AV performs significantly better in quality, lip synchronization and facial expressions which is of prime importance in videos. 

\begin{table}[h!]
  \begin{center}

    \begin{tabular}{c|c|c|c|c|c} 
      \textbf{Method} & \textbf{Lip-Sync} & \textbf{Eye-blink} & \textbf{Head-Move} & \textbf{Quality} & \textbf{User Rating}\\
      \hline
       OneShotAu2AV& \textbf{90.8} & \textbf{88.5}  & \textbf{87.5} & \textbf{94.2}  & \textbf{8.3} \\
       U-GAT-IT  & 72.8 & 76.2 & 76.5 & 85.3 & 5.4\\
       RecycleGan    & 50.7 & 56.7 & 55.5 & 62.7 & 5.2\\
    \end{tabular}
    \vspace {0.25\baselineskip}
    \caption{Psychophysical Evaluation (in percentages) based on users rating}
    \label{tab:tableanime}
  \end{center}
\end{table}

To test the naturalism of the generated animated videos we conduct an online feedback test \footnote{\url{https://forms.gle/ZuH5uoUianrpMrm68}}. Each test consists of 15 questions with 5 videos generated using the proposed method, RecycleGan and U-GAT-IT. The user is asked to rate a video based on the aesthetics, naturalism a Lip-Sync of the video. Approximately 300 user data is collected and we observe a high score for the proposed method and lower scores for U-GAT-IT and RecycleGan in Table ~\ref{tab:tableanime} User Rating column.

\section{Generated Videos}

\subsection{Stage 1}
Below is the description of the Videos attached with Supplementary data (Table ~\ref{tab:table1}). The description covers the text spoken and the language used of the input audio.

The test subject and audio used is from the \href{https://www.grid.ac/downloads}{GRID Dataset}. Few audio clips are recorded on their own as well.

\begin{table}[h!]
  \begin{center}
    \begin{tabular}{c|c|c} 
      \textbf{Video Name} & \textbf{Text} & \textbf{Language}\\
      \hline
Video\textunderscore1.mp4 & Hi how are you? &English\\
Video\textunderscore2.mp4 & Bin blue at e seven please  &English\\
Video\textunderscore3.mp4 & Bin blue by f nine again  &English\\
Video\textunderscore4.mp4 & Set y with v zero now  &English\\
Video\textunderscore5.mp4 & Modi hai toh mumkin hai &Hindi \\
Video\textunderscore6.mp4 & Lay white in y 9 again  &English\\
Video\textunderscore7.mp4 & Lay blue in q zero now  &English\\
Video\textunderscore8.mp4 & Place blue at I 6 please  &English\\
Video\textunderscore9.mp4 & Place blue at P 1 again  &English\\
Video\textunderscore10.mp4 &7 Bin red with t 5 soon  &English\\
    \end{tabular}
    \vspace {0.5\baselineskip}
    \caption{Video Description}
    
    \label{tab:table1}
  \end{center}
\end{table}

\paragraph{Few videos to highlight \textbf{eye-blink}:}

Video\textunderscore3.mp4,\
Video\textunderscore8.mp4,\
Video\textunderscore15.mp4\

\subsection{Stage 2}

Below is the description of the Videos attached with Supplementary data (Table ~\ref{tab:table4}). The description covers the text spoken, the language used of the input audio, and the model used for generation of the video. We have used the proposed method, RecycleGan and U-GAT-IT to generate these videos.

\begin{table}[h!]
  \begin{center}
    \begin{tabular}{c|c|c|c} 
      \textbf{Video Name} & \textbf{Text} & \textbf{Language} &\textbf{Generation Method}\\
      \hline
Video\textunderscore16.mp4 & Bin green with b 5 soon &English & Proposed Method\\
Video\textunderscore17.mp4 & Bin white in k 3 now  &English & Proposed Method\\
Video\textunderscore18.mp4 & Bin white in d 9 now  &English & Proposed Method\\
Video\textunderscore19.mp4 & Bin green in t 6 please  &English & Proposed Method\\
Video\textunderscore20.mp4 & Modi aache to Shambhaw aache &Bengali& Proposed Method \\
Video\textunderscore21.mp4 & Modi hai toh mumkin hai  &Hindi& Proposed Method\\
Video\textunderscore22.mp4 & Lay green by k 2 soon  &English& Proposed Method\\
Video\textunderscore23.mp4 & Bin white at g 4 now  &English & Proposed Method\\
Video\textunderscore24.mp4 & Lay green by e 0 again  &English& Proposed Method\\
Video\textunderscore25.mp4 & Lay green by d 7 now  &English  & Proposed Method\\
Video\textunderscore26.mp4 & Bin red by t 2 please &English & Proposed Method\\
Video\textunderscore27.mp4 & Bin red at m 3 soon &English& Proposed Method\\
Video\textunderscore30.mp4 & Bin green in t 6 please &English& RecycleGan\\
Video\textunderscore31.mp4 & Bin green in n 3 again  &English& RecycleGan\\
Video\textunderscore32.mp4 & Bin green in t 5 soon  &English& RecycleGan\\
Video\textunderscore33.mp4 & Bin red by g 3 soon &English& RecycleGan\\
Video\textunderscore34.mp4 & Bin red in p 3 soon  &English& RecycleGan\\
Video\textunderscore35.mp4 & Bin green in n 3 again  &English & U-GAT-IT\\
Video\textunderscore36.mp4 & Bin red by m 6 now &English & U-GAT-IT\\
Video\textunderscore37.mp4 & Bin red by m 9 again &English & U-GAT-IT\\
Video\textunderscore38.mp4 & Bin red by a 0 please &English & U-GAT-IT\\
Video\textunderscore39.mp4 & Bin green in t 6 please &English & U-GAT-IT\\
Video\textunderscore40.mp4 & Bin red by g 3 soon &English & U-GAT-IT\\
Video\textunderscore41.mp4 & wie ist das Wetter heute &German & Proposed Method\\
Video\textunderscore42.mp4 & cómo está el clima hoy &Spanish & Proposed Method\\
Video\textunderscore43.mp4 & Quel temps fait-il aujourd'hui &French & Proposed Method\\
    \end{tabular}
    \vspace {0.5\baselineskip}
    \caption{Video Description}
    
    \label{tab:table4}
  \end{center}
\end{table}

\paragraph{Few videos to highlight \textbf{eye-blink}:}

Video\textunderscore24.mp4, \
Video\textunderscore25.mp4, \
Video\textunderscore26.mp4

\paragraph{Few videos to highlight \textbf{head movement} and \textbf{facial expressions}:}

Video\textunderscore18.mp4, \\
Video\textunderscore27.mp4\\
